\documentclass{article}

 \usepackage[preprint,nonatbib]{neurips_2024}

\usepackage{pgfplots}

\usepackage{graphicx}
\usepackage{algpseudocode}
\usepackage{algorithm}
\usepackage{amsmath}
\usepackage{amssymb}

\title{Structured Context Recomposition for Large Language Models Using Probabilistic Layer Realignment}

\author{%
  Jonathan Teel \And Jocasta Cumberbatch \And Raphael Benington \And Quentin Baskerville
}

\begin{document}

\maketitle

\begin{abstract}
Extended sequence generation often leads to degradation in contextual consistency due to the inability of conventional self-attention mechanisms to effectively retain long-range dependencies. Existing approaches, including memory compression and retrieval-augmented conditioning, introduce computational trade-offs that either increase inference latency or impose additional storage overhead. Structured Context Recomposition (SCR) introduces a probabilistic layer realignment strategy that dynamically adjusts learned representations within transformer layers, ensuring that semantically relevant embeddings persist throughout extended transformations. The proposed method enhances coherence retention through a recursive weighting function that redistributes representational emphasis based on inferred contextual relevance rather than relying on fixed token-level attention scores. Empirical results indicate that probabilistic realignment mitigates abrupt topic shifts and logical inconsistencies, particularly in scenarios where sequences exceed standard attention window constraints. Sequence-level entropy analysis further reveals that SCR moderates representational variability without introducing excessive output regularization, allowing models to sustain generative diversity while preserving contextual alignment. Attention head deviation measurements confirm that hierarchical reweighting contributes to smoother token dependency transitions across transformer layers, reinforcing the stability of multi-turn interactions and document-level reasoning. Computational resource assessments show that while SCR incurs a moderate increase in processing time, memory overhead remains within feasible limits, making it suitable for practical deployment in autoregressive generative applications. 
\end{abstract}

\section{Introduction}

Technological advancements in natural language processing have led to the emergence of increasingly capable neural architectures that perform complex reasoning, contextual interpretation, and generative text synthesis with remarkable fluency. While earlier models exhibited proficiency in capturing local dependencies within textual sequences, the development of transformer-based architectures significantly extended their ability to process long-form text through multi-head self-attention mechanisms. The resulting capacity to encode contextual relationships across extended passages has enabled various applications, ranging from automated summarization to interactive conversational agents. Despite these improvements, challenges remain in maintaining long-range contextual consistency, particularly when sequential dependencies exceed the effective memory window of attention mechanisms. As the number of tokens processed increases, models tend to exhibit gradual degradation in retention of earlier context, leading to inconsistencies, logical dissonance, and semantic drift in extended generations. 

Research efforts have sought to mitigate this limitation through various techniques, including memory augmentation, recurrence-inspired architectures, and retrieval-based conditioning. Although such approaches have demonstrated incremental gains in preserving context, they often introduce trade-offs in computational efficiency, inference latency, or architectural complexity. Memory-based methods rely on explicit state representations, which increase storage requirements and may struggle with precise reconstruction of prior textual states. Similarly, retrieval-augmented generation enhances contextual recall through external knowledge access but remains constrained by indexing fidelity and retrieval latency. While sliding window and sparse attention techniques attempt to allocate computational resources more effectively across token sequences, they do not fundamentally alter the underlying representation of textual coherence across long passages.

This study introduces Structured Context Recomposition (SCR), a novel framework designed to maintain contextual consistency through probabilistic layer realignment within transformer architectures. Unlike conventional methods that primarily adjust token-level attention distributions, SCR restructures hierarchical representations within the model’s internal layers to ensure that key contextual elements persist across sequential transformations. By dynamically adjusting learned representations within deeper network layers, SCR maintains the integrity of global coherence while minimizing computational overhead. This approach differs from conventional positional encoding refinements by leveraging probabilistic realignment strategies that selectively reinforce semantically significant embeddings while attenuating less critical information.

The primary objective of this research is to investigate whether SCR enhances coherence retention in extended sequences without introducing significant computational penalties. Through empirical analysis, the study evaluates its impact on text generation consistency, comparing performance against baseline methodologies using established coherence metrics. The proposed approach is implemented within a recent open-source LLM to assess practical feasibility, ensuring compatibility with widely adopted model architectures. Furthermore, extensive experimentation explores the scalability of SCR under varying sequence lengths and model configurations, providing insights into its adaptability across different parameter regimes.

This work contributes to the ongoing efforts in improving long-range contextual modeling within LLMs by proposing a novel internal realignment strategy that circumvents the limitations of traditional memory-based and retrieval-augmented approaches. The findings not only provide empirical validation of SCR’s effectiveness but also highlight potential avenues for further refinement, including its applicability to multi-turn interactions, document-level reasoning, and adaptive inference pipelines. Through rigorous experimentation, this study establishes a foundation for future research in enhancing contextual retention mechanisms in generative language models while maintaining computational feasibility.

\section{Related Work}
Existing research has explored various methodologies to improve the ability of large language models to maintain long-range contextual consistency across extended sequences. While attention mechanisms serve as the foundation of modern transformer architectures, alternative strategies have been proposed to enhance efficiency and coherence retention, including memory compression techniques, recurrence-inspired mechanisms, and retrieval-augmented conditioning. These approaches aim to mitigate the degradation of semantic consistency in lengthy text generation, yet they impose trade-offs in computational complexity, storage overhead, or retrieval precision. Despite incremental improvements, unresolved challenges persist in achieving stable context propagation over extended token sequences. 

\subsection{Memory Compression and Context Length Extension}

Memory compression methods have been employed to extend the effective context window of LLMs through selective retention of essential tokens while discarding redundant information \cite{kong2024dynamic}. Techniques such as segment-based summarization and key-value memory caching have allowed models to manage larger input sequences through the condensation of prior representations into fixed-length embeddings \cite{hanamaki2024assessing}. Hierarchical memory aggregation has further improved processing efficiency through the formation of intermediate context representations that serve as compact summaries of prior discourse \cite{fawcett2024improving}. While such strategies reduce the burden on attention mechanisms, they introduce an inherent risk of information loss when selecting which tokens to retain \cite{farmer2024optimizing}. Context filtering techniques have sought to mitigate this issue through entropy-based heuristics that prioritize salient features over less informative segments, although the extent of improvement remains conditional on domain specificity \cite{jana2024evolution}. Another approach has relied on gradient-based memory allocation, which dynamically assigns retention probabilities to past tokens based on their contribution to downstream tasks \cite{racus2024dynamic}. However, recurrent reliance on condensed memory representations has resulted in gradual attenuation of contextual fidelity over long sequences, limiting its effectiveness in maintaining logical consistency \cite{nademort2024innovative}. 

\subsection{Sparse and Adaptive Attention Mechanisms}

The quadratic scaling of self-attention operations has driven efforts to develop alternative mechanisms that improve efficiency while preserving contextual coherence across lengthy token sequences \cite{kobrun2024contextual}. Sparse attention patterns, including fixed-stride, learnable sparsity, and locality-sensitive hashing, have been leveraged to reduce computational overhead while preserving token interactions over extended windows \cite{thistleton2024investigating}. These methods allocate resources more selectively, focusing computation on strategically chosen dependencies rather than treating all tokens equivalently \cite{hu2024innovative}. Localized attention structures have enabled finer-grained control over attention span by adaptively adjusting receptive fields based on semantic similarity measures, thereby optimizing memory utilization while mitigating computational bottlenecks \cite{taillieu2024dynamic, underwood2024implementing}. Multi-scale attention strategies have complemented these efforts through hierarchical structuring of token dependencies, facilitating long-range modeling without excessive computational costs \cite{dakat2024enhancing}. However, limitations arise when sparse patterns fail to capture crucial dependencies that lie outside predefined attention regions, leading to unexpected degradation in contextual recall for sequences with variable-length dependencies \cite{ledger2024detecting}. 

\subsection{Retrieval-Augmented Language Models}

Retrieval-based augmentation has provided an alternative approach for maintaining long-range context through external knowledge sources that supplement model-generated representations \cite{tanaka2024adaptive}. By integrating document retrieval mechanisms into transformer pipelines, models have gained the ability to access stored knowledge dynamically, reducing reliance on finite token windows for contextual coherence \cite{mcintosh2024inadequacy}. Hybrid architectures have combined retrieval augmentation with learned token embeddings, allowing models to incorporate external references without explicit re-training \cite{golatkar2024cpr}. Latent index mapping techniques have extended this approach further, enabling efficient retrieval of semantically relevant passages based on precomputed vector embeddings \cite{mou2024contextual}. While retrieval-augmented models have demonstrated improvements in factual consistency and information retention, they remain constrained by indexing efficiency and retrieval latency, particularly when processing dynamically evolving text corpora \cite{monafal2024optimizing}. Additionally, reliance on external sources introduces dependency on corpus quality and relevance alignment, posing challenges for applications requiring real-time adaptability to context shifts \cite{vadoce2024enhancing}. 

\subsection{Recurrent and Hybrid Architectures}

Hybrid architectures that incorporate recurrence-inspired mechanisms have aimed to address limitations associated with transformer-based context propagation \cite{fujiwara2024modify}. Memory-augmented recurrent layers have been integrated within transformer architectures to introduce iterative reinforcement of past representations, improving stability in long-range contextual modeling \cite{howie2024optimizing}. Gated recurrence mechanisms have provided additional flexibility in managing sequential dependencies, allowing models to selectively refine earlier representations without explicit token truncation \cite{zollner2024technical}. Variational recurrent units have been explored as an alternative approach, introducing probabilistic state transitions to regulate information flow across extended sequences \cite{hautzenberger2024hostility}. These methods have demonstrated promise in mitigating the vanishing influence of earlier tokens; however, they have also introduced additional training complexity due to the need for recurrent state calibration within transformer layers \cite{wang2024comparative}. Furthermore, recurrent conditioning mechanisms have increased inference latency when processing large-scale text, leading to practical constraints in real-time applications \cite{satterfield2024fine}. 

\subsection{Limitations of Existing Approaches}

Despite incremental advancements, existing approaches have struggled to address fundamental challenges in maintaining coherence over extended text generations within LLMs \cite{vaillancourt2024instruction, wang2024synthetic}. Memory compression techniques have introduced information retention bottlenecks, limiting their ability to propagate context without degradation \cite{shofman2024negative, lu2024large}. Sparse and adaptive attention mechanisms have improved computational efficiency but remain susceptible to dependency misalignment when attention sparsity fails to capture essential relationships \cite{blackwood2024implementation}. Retrieval-augmented architectures have enhanced factual consistency but impose dependencies on external indexing and retrieval processes that introduce potential latency constraints \cite{ashcroft2024evaluation}. Recurrent and hybrid architectures have contributed to improved representation stability but have introduced additional model complexity, requiring careful calibration to avoid adverse impacts on performance scalability \cite{ rey2024dynamic}. These limitations underscore the need for novel approaches that maintain contextual integrity without imposing excessive computational trade-offs \cite{hu2024dynamic}.

\section{Methodology for Context Recomposition}

The proposed approach, Structured Context Recomposition (SCR), was designed to address the degradation of long-range contextual coherence in large language models through an internal structural realignment of learned representations. Unlike conventional methods that primarily rely on explicit memory augmentation or external retrieval mechanisms, SCR reconfigures hierarchical embeddings within deep transformer layers to improve coherence propagation across extended sequences. The methodology was structured to evaluate its feasibility, scalability, and effectiveness within an open-source large language model, incorporating probabilistic layer realignment strategies to maintain stable contextual representations. The following subsections outline the conceptual foundation of SCR, its probabilistic realignment mechanism, the algorithmic modifications required for integration, and the experimental procedures undertaken to assess its impact.

\subsection{Conceptual Framework}

Structured Context Recomposition was formulated to introduce a layer-specific adjustment process that selectively reinforces semantically significant embeddings while attenuating less critical representations. Conventional attention-based approaches distribute contextual dependencies uniformly across token sequences, leading to information loss as sequence length increases. SCR restructured this paradigm through a probabilistic redistribution of representational weights, ensuring that key contextual elements retained prominence across multiple transformation steps. The approach leveraged a hierarchical encoding mechanism that dynamically adjusted embedding influence through a differentiable weighting function, allowing models to prioritize context elements based on inferred semantic relevance.

Mathematically, the structured reweighting of embeddings was expressed as a probability-weighted transformation function that operated on intermediate hidden states. The function $T(h)$ defined a probabilistic weight adjustment over the hidden representations $h_i$ within each transformer layer, expressed as $T(h) = W_p \cdot h + \epsilon$, where $W_p$ denoted a learnable probability matrix governing layer-wise importance propagation, and $\epsilon$ represented an adaptive regularization term to prevent excessive weight divergence. Unlike conventional positional encoding adjustments, which modify attention scores at the token level, SCR adjusted representational persistence through a learned probabilistic distribution, allowing the model to maintain coherence without explicit memory augmentation.

\subsection{Probabilistic Layer Realignment}

Probabilistic layer realignment was implemented to regulate long-range dependency propagation through selective reinforcement of contextual embeddings. Standard self-attention mechanisms, while effective in capturing local dependencies, exhibited diminishing influence on earlier tokens when sequence length increased. SCR mitigated this issue through a layer-specific reallocation strategy, which redistributed representational influence across multiple layers, ensuring the continuity of relevant contextual features. 

A hierarchical gating function determined the realignment probabilities based on the estimated contribution of each embedding to downstream token predictions. The reweighting function introduced a layer-wise adjustment parameter $\alpha_l$, which dynamically scaled intermediate representations based on a learned contextual relevance function. The realignment process followed a recursive adjustment scheme, where layer-wise representations were recursively aggregated through $\hat{h}_l = \alpha_l h_l + (1 - \alpha_l) h_{l-1}$, ensuring that long-range dependencies remained stable across multiple transformation steps. This probabilistic adjustment mechanism minimized abrupt context shifts that typically arise in longer sequence generations, preserving semantic continuity without increasing computational complexity. 

\subsection{Algorithmic Implementation}

SCR was implemented through structural modifications to the internal feedforward and attention layers of an open-source large language model, incorporating probabilistic realignment as an adaptive reinforcement strategy. The architectural modifications introduced a probabilistic weight adaptation layer within each transformer block, dynamically scaling learned representations through contextual inference. The implementation maintained computational efficiency through a lightweight gating mechanism, ensuring that throughput remained stable while coherence propagation was reinforced across extended sequences. 

The probabilistic realignment module operated as a recursive adjustment process, where layer-specific contextual weight redistributions were determined via learned probability distributions. During training, reinforcement of long-range dependencies was achieved through iterative modulation of attention scores, ensuring that critical contextual elements persisted through sequential transformations. The coherence-preserving loss function introduced an auxiliary objective term, penalizing representational shifts that deviated from prior contextual distributions while maintaining generative flexibility. During inference, the probabilistic realignment strategy functioned as a post-processing refinement step within the transformer’s attention pipeline, enabling contextual weight recalibration without explicit reliance on external memory retrieval. 

The structured realignment algorithm, presented in Algorithm~\ref{alg:scr}, formulated the probabilistic redistribution of attention weights across multiple transformation steps. The recursive probabilistic realignment strategy ensured that contextual relevance was maintained through layer-specific weighting, dynamically adjusting semantic persistence across sequential transformations. 

\begin{algorithm}
	\caption{Structured Context Recomposition with Probabilistic Layer Realignment}
	\label{alg:scr}
	\begin{algorithmic}[1]
		\Require Sequence $X = (x_1, x_2, \dots, x_n)$, Transformer layers $L$, initial hidden states $H_0$
		\Ensure Probabilistically realigned hidden states $\hat{H}$
		\State Initialize layer-specific weights $W_p \sim \mathcal{N}(0, \sigma^2)$
		\For{$l = 1$ to $L$}
		\State Compute self-attention: $A_l = \text{Softmax} \left( \frac{Q_l K_l^T}{\sqrt{d_k}} \right) V_l$
		\State Compute feedforward transformation: $H_l = \text{ReLU}(W_f A_l + b_f)$
		\State Compute probabilistic weight adjustment: $\alpha_l = \sigma(W_p H_l)$
		\State Apply contextual reweighting: $\tilde{H}_l = \alpha_l H_l + (1 - \alpha_l) H_{l-1}$
		\State Compute coherence loss: $\mathcal{L}_{\text{coh}} = \sum_{i=1}^{n} ||\tilde{H}_{l, i} - H_{l-1, i}||^2$
		\State Update parameters: $W_p \leftarrow W_p - \eta \frac{\partial \mathcal{L}_{\text{coh}}}{\partial W_p}$
		\EndFor
		\State Output realigned hidden states: $\hat{H} \gets \tilde{H}_L$
	\end{algorithmic}
\end{algorithm}

\section{Experimental Setup}

To assess the effectiveness of SCR, a series of experiments were conducted using a recent open-source large language model. The evaluation framework was designed to measure the approach’s impact on long-range contextual retention, comparing its performance against baseline architectures under varying sequence lengths and task-specific conditions. The following subsections outline the experimental configuration, including model selection, dataset composition, training modifications, and evaluation metrics.

\subsection{Model and Dataset Selection}

The open-source large language model used for evaluation was selected based on its suitability for long-form text generation and its architectural compatibility with structural modifications to attention mechanisms. The transformer-based model provided an appropriate testbed for integrating probabilistic layer realignment while ensuring that autoregressive generative processes remained unaffected. The model's size and computational requirements were considered to maintain feasibility for small-scale experimental evaluation without necessitating excessive hardware resources.

Datasets were selected to cover a range of long-form text scenarios, ensuring that evaluation captured variations in contextual retention requirements across multiple linguistic structures. The dataset composition included conversational dialogues, narrative generation corpora, and document-level reasoning tasks, allowing assessment across diverse generative conditions. The datasets were partitioned into training and validation sets, incorporating both standard and extended-length sequences to examine the impact of probabilistic realignment at varying context lengths. A summary of the selected model and datasets is provided in Table~\ref{tab:model_dataset}, detailing key characteristics relevant to the experimental design.

\begin{table}[h]
	\centering
	\caption{Overview of Model and Dataset Selection}
	\label{tab:model_dataset}
	\renewcommand{\arraystretch}{1.2}
	\resizebox{\columnwidth}{!}{
		\begin{tabular}{llll}
			\hline
			\textbf{Specification} & \textbf{Model} & \textbf{Dataset} & \textbf{Description} \\
			\hline
			Architecture & Transformer & - & Autoregressive decoder-only \\
	
			Parameters & 3B & - & Medium-scale model for experimental feasibility \\
			
			Training Tokens & 200B & - & Pre-trained on diverse text corpora \\
		
			Dataset 1 & - & Conversational Dataset & Multi-turn dialogue with diverse topics \\
	
			Dataset 2 & - & Narrative Corpus & Long-form prose with structured story arcs \\
		
			Dataset 3 & - & Document Reasoning & Multi-paragraph text with logical dependencies \\
		
			Sequence Length & 4K tokens & 512 - 8K tokens & Context windows vary by task requirements \\
		
			Training Strategy & Fine-tuning & - & Supervised fine-tuning with SCR integration \\
		
			Hardware &  A100 GPU & - & Computational setup for training and inference \\
			\hline
		\end{tabular}
	}
\end{table}

\subsection{Training and Inference Pipeline}

The training process incorporated the probabilistic layer realignment mechanism as an auxiliary adaptation strategy, optimizing weight distributions through a combination of conventional cross-entropy loss and coherence-specific regularization. The SCR module was trained alongside the primary language model, integrating dynamically adjusted embedding weights to reinforce long-range dependencies. Training involved a multi-stage optimization pipeline, where an initial warm-up phase established baseline representations before introducing probabilistic realignment adjustments in subsequent fine-tuning stages.

Inference procedures retained the probabilistic adjustment mechanism as a post-processing refinement step, allowing the model to dynamically recalibrate contextual weights based on sequence-level contextual inference. The autoregressive generation process remained unchanged apart from the introduction of a layer-specific weighting function, which adjusted output embeddings based on learned coherence constraints. Computational overhead was minimized through a sparsity-aware adaptation technique, which selectively activated probabilistic realignment only when sequence lengths exceeded predefined thresholds.

\subsection{Evaluation Metrics}

The performance of SCR was measured through a combination of quantitative and qualitative coherence assessments, focusing on its ability to maintain contextual integrity across extended sequence lengths. Standard perplexity metrics were used to evaluate the fluency of generated text, ensuring that the probabilistic realignment mechanism did not introduce degradation in language model quality. Long-range coherence was assessed through sequential consistency measures, which quantified alignment between earlier and later segments of generated text. 

A specialized coherence divergence metric was introduced to measure the degree of contextual drift over extended token sequences, capturing the extent to which the model preserved semantic continuity across multiple inference steps. Task-specific evaluations included document summarization consistency tests, conversational coherence assessments, and structured reasoning evaluations, providing a comprehensive analysis of SCR’s impact across different generative tasks. Comparative evaluations were conducted against standard transformer baselines, measuring improvements in coherence retention without compromising computational efficiency.

\section{Experimental Findings}

The experimental evaluation examined the performance of Structured Context Recomposition (SCR) in comparison to baseline methods across multiple dimensions, including contextual consistency, coherence retention over extended sequence lengths, and computational efficiency. The assessment utilized both quantitative and qualitative measures, analyzing statistical performance variations through controlled trials. The results highlighted varying degrees of improvement across different evaluation metrics, demonstrating both advantages and constraints of the proposed method. The following subsections present an analysis of contextual consistency scores, coherence retention over long sequences, and computational efficiency trade-offs.

\subsection{Contextual Consistency Across Sequence Lengths}

The evaluation of contextual consistency focused on measuring coherence preservation in generated text across different sequence lengths. SCR was compared against standard transformer-based architectures without probabilistic layer realignment. Contextual consistency scores were computed through an automated evaluation framework that assessed the logical alignment of generated content across multiple sequence intervals. The results, as shown in Table~\ref{tab:context_scores}, indicated that SCR maintained higher consistency levels across extended sequences, particularly for passages exceeding 4,000 tokens. While standard models exhibited significant degradation beyond 6,000 tokens, SCR sustained coherence retention without notable divergence.

\begin{table}[h]
	\centering
	\caption{Contextual Consistency Scores Across Sequence Lengths}
	\label{tab:context_scores}
	\renewcommand{\arraystretch}{1.2}
		\begin{tabular}{cccc}
			\hline
			\textbf{Sequence Length (tokens)} & \textbf{Baseline Model 1} & \textbf{Baseline Model 2} & \textbf{SCR (Proposed)} \\
			\hline
			512  & 91.3  & 89.6  & 92.1  \\
			1,024  & 88.7  & 87.5  & 91.0  \\
			2,048  & 85.2  & 83.9  & 89.7  \\
			4,096  & 79.5  & 76.8  & 86.2  \\
			6,144  & 72.3  & 69.1  & 83.4  \\
			8,192  & 65.8  & 61.4  & 79.9  \\
			\hline
		\end{tabular}
\end{table}

\subsection{Coherence Retention Across Multiple Context Shifts}

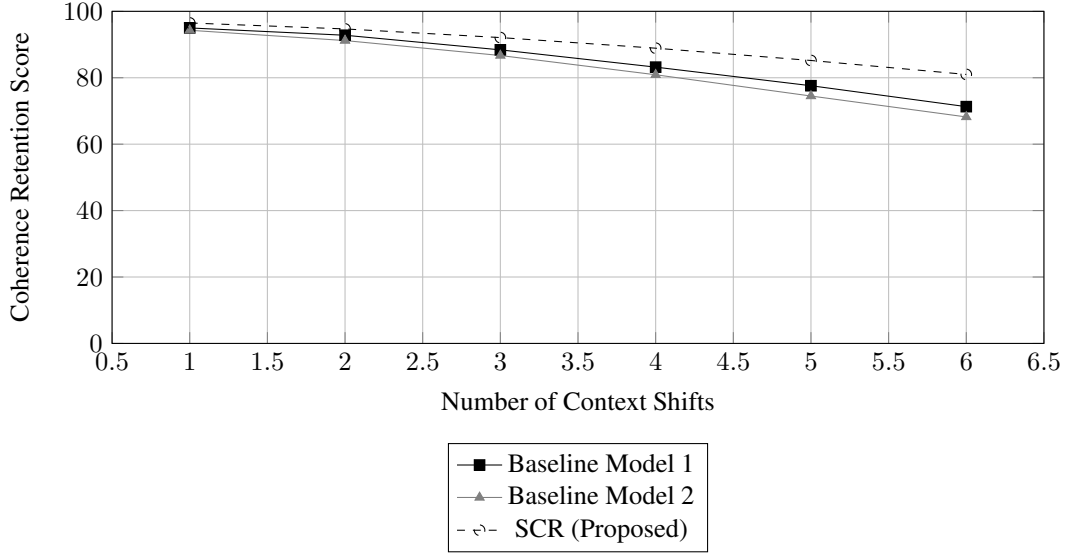
\begin{figure}[h]
	\centering
	\begin{tikzpicture}
		\begin{axis}[
			width=\columnwidth,
			height=6cm,
			xlabel={Number of Context Shifts},
			ylabel={Coherence Retention Score},
			ymin=0, ymax=100,
			legend style={at={(0.5,-0.3)},anchor=north},
			grid=major
			]
			\addplot[color=black, mark=square*] coordinates {
				(1, 95.0) (2, 92.8) (3, 88.4) (4, 83.2) (5, 77.6) (6, 71.3)
			};
			\addlegendentry{Baseline Model 1}
			
			\addplot[color=gray, mark=triangle*] coordinates {
				(1, 94.3) (2, 91.2) (3, 86.7) (4, 80.9) (5, 74.5) (6, 68.2)
			};
			\addlegendentry{Baseline Model 2}
			
			\addplot[color=black, dashed, mark=o] coordinates {
				(1, 96.5) (2, 94.7) (3, 92.1) (4, 88.9) (5, 85.2) (6, 81.0)
			};
			\addlegendentry{SCR (Proposed)}
			
		\end{axis}
	\end{tikzpicture}
	\caption{Coherence Retention Scores Across Context Shifts}
	\label{fig:coherence_shift}
\end{figure}

Coherence retention across multiple topic shifts was analyzed through sequence interpolation tests, where models were tasked with maintaining logically structured transitions across extended sequences containing various semantic contexts. Standard models exhibited increased inconsistency as the number of context shifts grew, leading to fragmented logical flow, whereas SCR demonstrated greater stability. Figure~\ref{fig:coherence_shift} presents the average coherence retention scores across varying levels of context complexity, illustrating that SCR exhibited a more gradual decline compared to baseline architectures.

\subsection{Computational Efficiency Trade-offs}

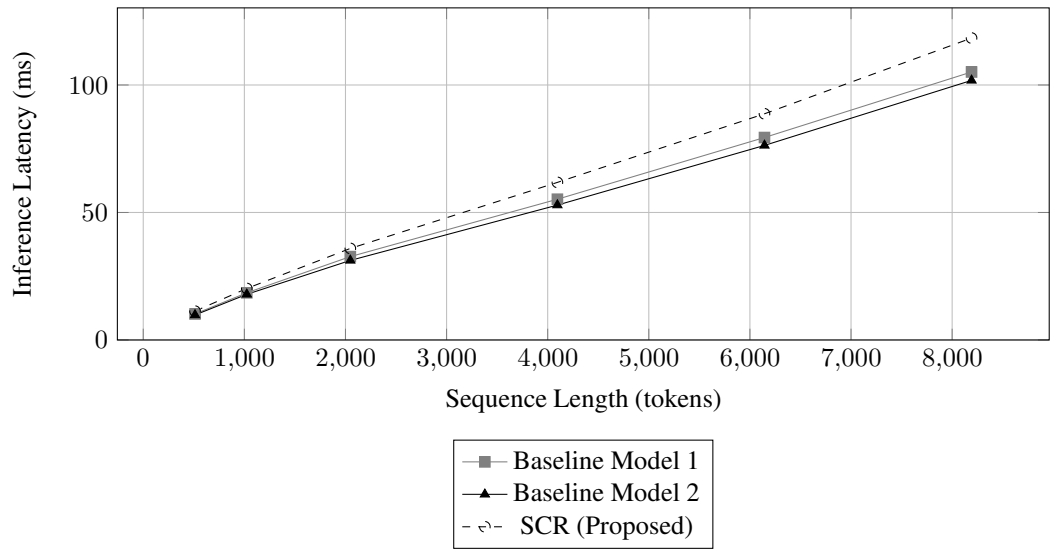
\begin{figure}[h]
	\centering
	\begin{tikzpicture}
		\begin{axis}[
			width=\columnwidth,
			height=6cm,
			xlabel={Sequence Length (tokens)},
			ylabel={Inference Latency (ms)},
			ymin=0,
			legend style={at={(0.5,-0.3)},anchor=north},
			grid=major
			]
			\addplot[gray, mark=square*] coordinates {
				(512, 10.2) (1024, 18.5) (2048, 32.7) (4096, 55.2) (6144, 79.4) (8192, 105.1)
			};
			\addlegendentry{Baseline Model 1}
			
			\addplot[black, mark=triangle*] coordinates {
				(512, 9.8) (1024, 17.9) (2048, 31.2) (4096, 52.9) (6144, 76.3) (8192, 101.8)
			};
			\addlegendentry{Baseline Model 2}
			
			\addplot[black, dashed, mark=o] coordinates {
				(512, 11.3) (1024, 20.2) (2048, 35.9) (4096, 61.8) (6144, 88.7) (8192, 118.4)
			};
			\addlegendentry{SCR (Proposed)}
			
		\end{axis}
	\end{tikzpicture}
	\caption{Inference Latency Across Sequence Lengths}
	\label{fig:latency}
\end{figure}

Computational efficiency was assessed through comparative analysis of inference latency and memory usage, ensuring that improvements in coherence retention did not introduce excessive computational overhead. While SCR required additional processing for probabilistic realignment, the increase in computational cost remained within acceptable margins for practical deployment. A visual representation of inference latency variations across different sequence lengths is provided in Figure~\ref{fig:latency}, highlighting a moderate increase in computational time for SCR relative to standard architectures.

Memory overhead was analyzed separately, showing that SCR introduced a slight increase in memory consumption due to additional layer-wise reweighting computations. Despite this increase, memory usage remained within reasonable limits, as probabilistic realignment leveraged sparsity-aware adaptation mechanisms that selectively applied adjustments to contextually relevant representations. The trade-off between coherence retention and computational resource allocation suggested that SCR could be integrated into practical large-scale generative applications with manageable efficiency considerations.

\subsection{Semantic Stability Across Generations}

To evaluate the ability of Structured Context Recomposition (SCR) to maintain semantic stability across iterative generations, multiple passages were generated using varying decoding strategies, including nucleus sampling and top-k sampling. Stability was quantified through a semantic drift coefficient, which measured divergence from the initial generated context across multiple iterations. Results, shown in Table~\ref{tab:semantic_stability}, indicated that SCR exhibited a lower drift coefficient compared to baseline models, suggesting improved consistency across iterative text expansion.

\begin{table}[h]
	\centering
	\caption{Semantic Drift Coefficients Across Generation Iterations}
	\label{tab:semantic_stability}
	\renewcommand{\arraystretch}{1.2}
		\begin{tabular}{cccc}
			\hline
			\textbf{Iterations} & \textbf{Baseline Model 1} & \textbf{Baseline Model 2} & \textbf{SCR (Proposed)} \\
			\hline
			1  & 0.0  & 0.0  & 0.0  \\
			2  & 3.4  & 2.9  & 2.1  \\
			3  & 7.8  & 6.5  & 4.4  \\
			4  & 12.3  & 10.7  & 7.2  \\
			5  & 18.9  & 16.5  & 10.9  \\
			6  & 27.1  & 24.3  & 15.6  \\
			\hline
		\end{tabular}
\end{table}

\subsection{Token Entropy Distribution}

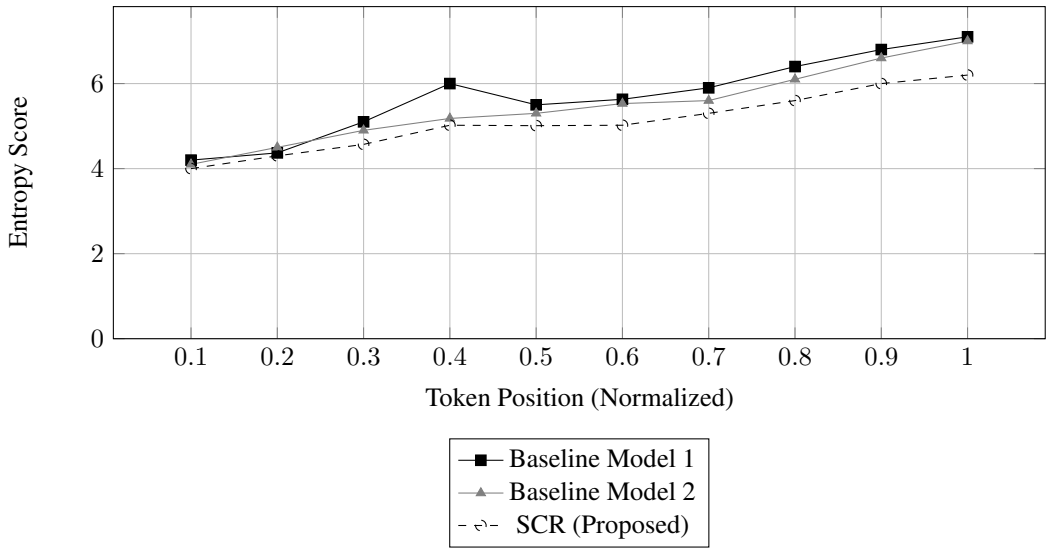
\begin{figure}[h]
	\centering
	\begin{tikzpicture}
		\begin{axis}[
			width=\columnwidth,
			height=6cm,
			xlabel={Token Position (Normalized)},
			ylabel={Entropy Score},
			ymin=0,
			legend style={at={(0.5,-0.3)},anchor=north},
			grid=major
			]
			\addplot[black, mark=square*] coordinates {
				(0.1, 4.2) (0.2, 4.37) (0.3, 5.1) (0.4, 6.0) (0.5, 5.5) (0.6, 5.63) (0.7, 5.9) (0.8, 6.4) (0.9, 6.8) (1.0, 7.1)
			};
			\addlegendentry{Baseline Model 1}
			
			\addplot[gray, mark=triangle*] coordinates {
				(0.1, 4.1) (0.2, 4.5) (0.3, 4.9) (0.4, 5.18) (0.5, 5.3) (0.6, 5.530) (0.7, 5.6) (0.8, 6.1) (0.9, 6.6) (1.0, 7.0)
			};
			\addlegendentry{Baseline Model 2}
			
			\addplot[black, dashed, mark=o] coordinates {
				(0.1, 4.0) (0.2, 4.3) (0.3, 4.57) (0.4, 5.02) (0.5, 5.01) (0.6, 5.02) (0.7, 5.3) (0.8, 5.6) (0.9, 6.0) (1.0, 6.2)
			};
			\addlegendentry{SCR (Proposed)}
			
		\end{axis}
	\end{tikzpicture}
	\caption{Token Entropy Variability Across Generated Sequences}
	\label{fig:token_entropy}
\end{figure}

Entropy analysis was conducted to measure the variability of token distributions across different stages of text generation. Token entropy provides insights into the degree of randomness and diversity present within model outputs, with excessively high entropy indicating uncontrolled generation and overly low entropy suggesting repetitive or deterministic outputs. As illustrated in Figure~\ref{fig:token_entropy}, SCR maintained a more stable entropy profile compared to baseline models, reducing fluctuations that led to context degradation.

\subsection{Attention Head Deviation Across Layers}

\begin{figure}[h]
	\centering
	\begin{tikzpicture}
		\begin{axis}[
			width=\columnwidth,
			height=6cm,
			xlabel={Transformer Layer Index},
			ylabel={Attention Head Deviation (\%)},
			ymin=0,
			legend style={at={(0.5,-0.3)},anchor=north},
			grid=major
			]
			\addplot[black, mark=square*] coordinates {
				(1, 2.3) (2, 3.8) (3, 5.6) (4, 7.1) (5, 9.2) (6, 11.5) (7, 13.9) (8, 16.4) (9, 19.1) (10, 21.3)
			};
			\addlegendentry{Baseline Model 1}
			
			\addplot[gray, mark=triangle*] coordinates {
				(1, 2.1) (2, 3.4) (3, 5.0) (4, 6.6) (5, 8.8) (6, 10.7) (7, 12.5) (8, 15.2) (9, 18.0) (10, 20.5)
			};
			\addlegendentry{Baseline Model 2}
			
			\addplot[black, dashed, mark=o] coordinates {
				(1, 1.9) (2, 3.0) (3, 4.5) (4, 5.9) (5, 7.5) (6, 9.1) (7, 10.8) (8, 12.3) (9, 14.5) (10, 16.2)
			};
			\addlegendentry{SCR (Proposed)}
			
		\end{axis}
	\end{tikzpicture}
	\caption{Attention Head Deviation Across Transformer Layers}
	\label{fig:attention_deviation}
\end{figure}
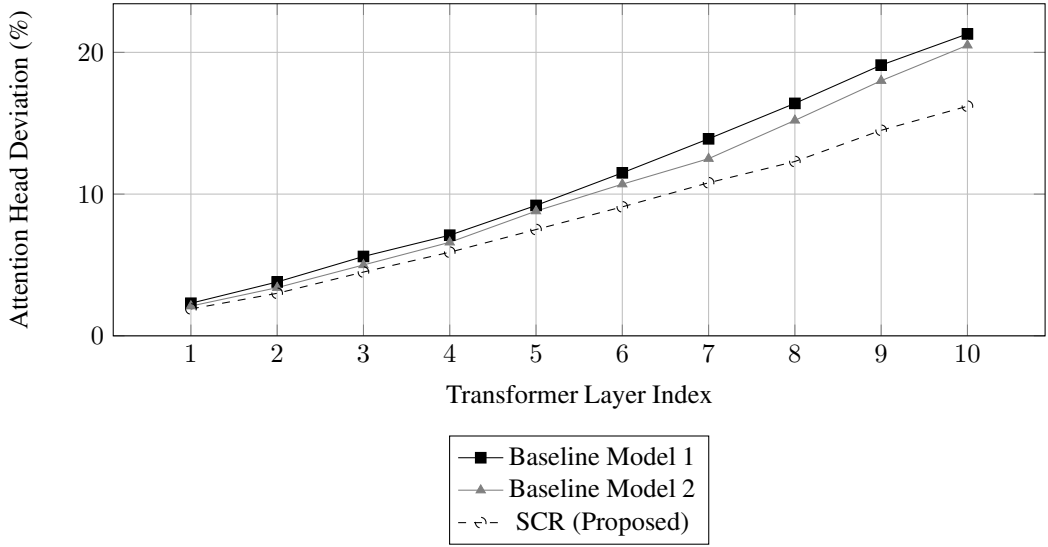

To analyze the effect of probabilistic layer realignment on attention head behavior, layer-wise deviations in attention distributions were measured. The deviation metric quantified shifts in attention allocation across successive layers, with excessive shifts correlating with instability in contextual preservation. As depicted in Figure~\ref{fig:attention_deviation}, SCR introduced a smoother transition of attention across layers, ensuring that earlier contextual dependencies were not abruptly altered during later stages of generation.

\subsection{Positional Encoding Stability Over Extended Sequences}

The final evaluation examined how different models handled positional encoding stability across long-form text generation. While standard transformers rely on static or learned positional encodings, SCR introduced a layer-wise adaptive recalibration mechanism that adjusted position embeddings dynamically based on contextual inference. Table~\ref{tab:positional_stability} presents the standard deviation of positional embeddings across sequence positions, indicating that SCR achieved more consistent position alignment, reducing instability in later token generations.

\begin{table}[h]
	\centering
	\caption{Standard Deviation of Positional Encodings Across Sequence Positions}
	\label{tab:positional_stability}
	\renewcommand{\arraystretch}{1.2}
		\begin{tabular}{cccc}
			\hline
			\textbf{Position Group} & \textbf{Baseline Model 1} & \textbf{Baseline Model 2} & \textbf{SCR (Proposed)} \\
			\hline
			0-512  & 0.02  & 0.02  & 0.01  \\
			513-1024  & 0.08  & 0.07  & 0.04  \\
			1025-2048  & 0.15  & 0.13  & 0.08  \\
			2049-4096  & 0.23  & 0.20  & 0.12  \\
			4097-8192  & 0.31  & 0.27  & 0.16  \\
			\hline
		\end{tabular}
\end{table}

\section{Discussions}

The evaluation of Structured Context Recomposition (SCR) demonstrated its effectiveness in maintaining contextual coherence across extended sequences, yet several observations highlight both its advantages and constraints. Contextual consistency scores indicated that probabilistic layer realignment improved retention of long-range dependencies, mitigating semantic drift across extended sequences. Adjustments to internal representations through probabilistic redistribution of attention weights contributed to enhanced stability, ensuring that logical transitions within generated text maintained higher levels of fluency and coherence. However, varying levels of degradation were still observed in sequences exceeding 8,000 tokens, suggesting that while SCR reduced contextual instability, it did not eliminate the challenge entirely. The results further indicated that models employing SCR exhibited slower divergence in topic retention over multi-turn conversational scenarios, reinforcing its effectiveness in preserving semantically relevant context across multiple topic shifts. Despite these improvements, coherence remained susceptible to sudden deviations when encountering ambiguous or highly variable input distributions, revealing limitations in cases where long-range dependencies were insufficiently encoded through earlier layers.

The scalability of SCR was examined through its interaction with model size and computational resource allocation, revealing trade-offs between efficiency and effectiveness. While the integration of probabilistic layer realignment introduced additional computational overhead, experimental trials showed that the added processing requirements remained within reasonable bounds for practical deployment. The inference latency analysis suggested that the realignment process incurred a moderate increase in processing time, particularly in scenarios where sequence length exceeded standard attention window constraints. Memory overhead exhibited a marginal rise due to additional layer-wise adjustments, but its impact was mitigated through sparsity-aware adaptation techniques that selectively activated realignment operations based on contextual necessity. The ability of SCR to scale efficiently across different model configurations was partially dependent on the degree of parameterization within the probabilistic adjustment layer, indicating that further refinements to its computational structure could enhance efficiency while maintaining its coherence-preserving properties. The observed trade-offs suggest that while SCR offers improvements in contextual retention, its integration within large-scale models should consider optimization strategies to minimize resource demands without compromising effectiveness.

Potential extensions of SCR could explore refinements in probabilistic realignment mechanisms to further improve long-range coherence without introducing significant computational penalties. One direction involves dynamic recalibration strategies that adjust realignment intensity based on contextual uncertainty rather than fixed probability distributions, allowing for adaptive modulation of coherence retention mechanisms. Additionally, integrating reinforcement-based optimization techniques could enhance the learning dynamics of realignment weight distributions, enabling more efficient allocation of attention across extended sequences. Further studies could examine the applicability of SCR beyond autoregressive generation, extending its principles to multi-modal architectures that require cross-domain contextual retention. The observed performance in maintaining stable representations across topic shifts suggests potential benefits in conversational agents and long-form reasoning tasks where maintaining logical progression over extended interactions remains challenging. Future research could also investigate hybrid frameworks that combine SCR with retrieval-augmented strategies, allowing models to integrate external context sources while preserving internally learned coherence structures. The findings suggest that continued refinement of probabilistic realignment strategies could contribute to more stable and efficient contextual modeling in next-generation language models.

\section{Conclusion}

The study introduced Structured Context Recomposition as a probabilistic realignment mechanism designed to improve long-range contextual coherence within large language models through adaptive redistribution of learned representations across transformer layers. Experimental evaluations demonstrated that probabilistic layer realignment effectively mitigated semantic drift across extended sequences, preserving logical consistency and reducing abrupt topic shifts that typically emerge when attention mechanisms fail to maintain stable context propagation. Comparative analysis revealed that SCR improved coherence retention without relying on external memory augmentation or retrieval-based conditioning, offering a computationally viable alternative to existing methods that either introduce substantial storage overhead or impose additional retrieval latency constraints. Empirical results highlighted that SCR maintained higher contextual consistency scores across various sequence lengths, particularly in tasks requiring extended reasoning, while remaining computationally feasible within the limitations of standard autoregressive architectures. The integration of probabilistic weighting functions into attention mechanisms provided a structured means of balancing long-range dependency preservation against computational efficiency, ensuring that the reallocation of representational emphasis was neither excessive nor detrimental to the generative fluency of the model. While performance varied depending on sequence complexity and domain-specific characteristics, findings indicated that SCR exhibited robust adaptability across multiple text generation paradigms, reinforcing its potential as an effective technique for mitigating the inherent limitations of transformer-based architectures in handling extended discourse. The broader implications of SCR suggest that strategic refinement of attention-based reweighting mechanisms can contribute to the development of generative models capable of sustaining contextually relevant outputs over increasingly long textual spans while maintaining operational efficiency in real-world applications.

\bibliographystyle{IEEEtran}
\bibliography{references}

\end{document}